\documentclass[10pt,twocolumn,letterpaper]{article}

\usepackage{iccv}
\usepackage{times}
\usepackage{epsfig}
\usepackage{graphicx}
\usepackage{amsmath}
\usepackage{amssymb}

\usepackage{tabularx,ragged2e}
\usepackage{booktabs,calc}
\usepackage{multirow}
\usepackage{hhline}
\usepackage{array}
\usepackage{float}
\usepackage{makecell}
\usepackage{xcolor}
\usepackage{mathtools}
\usepackage{enumitem}
\usepackage{cite}
\usepackage{dsfont}
\usepackage{pifont}% http://ctan.org/pkg/pifont
\usepackage{balance}

\newcolumntype{Y}{>{\centering\arraybackslash}X}
\newcolumntype{s}{>{\hsize=.3\hsize}Y}
\newcolumntype{t}{>{\hsize=.2\hsize}Y}
\newcolumntype{d}{>{\centering\hsize=.3\hsize}X}

\DeclareMathOperator*{\argmin}{argmin}

\usepackage[pagebackref=true,breaklinks=true,letterpaper=true,colorlinks,bookmarks=false]{hyperref}

\iccvfinalcopy % *** Uncomment this line for the final submission

\def\iccvPaperID{1785} % *** Enter the ICCV Paper ID here

\ificcvfinal\fi

\newcommand{\V}[1]{\mathbf{#1}}

\begin{document}

%%%%%%%%% TITLE
\title{\vspace{-6mm}KAMA: 3D Keypoint Aware Body Mesh Articulation\vspace{-6mm}}
\author{Umar Iqbal~~~~~~Kevin Xie~~~~~~Yunrong Guo~~~~~~Jan Kautz~~~~~~Pavlo Molchanov\\
NVIDIA \\
{\tt\small\{uiqbal,chxie,kellyg,pmolchanov,jkautz\}@nvidia.com}
}

\maketitle

\begin{abstract}
We present KAMA, a 3D Keypoint Aware Mesh Articulation approach that allows us to estimate a human body mesh from the positions of 3D body keypoints. To this end, we learn to estimate 3D positions of 26 body keypoints and propose an analytical solution to articulate a parametric body model, SMPL, via a set of straightforward geometric transformations. 
Since keypoint estimation directly relies on image clues, our approach offers significantly better alignment to image content when compared to state-of-the-art approaches. Our proposed approach does not require any paired mesh annotations and is able to achieve state-of-the-art mesh fittings through 3D keypoint regression only. 
Results on the challenging 3DPW and Human3.6M demonstrate that our approach yields state-of-the-art body mesh fittings. 
\end{abstract}

\vspace{-2mm}
\section{Introduction}

\begin{figure}
  \centering
    \setlength{\tabcolsep}{1pt}
    \renewcommand{\arraystretch}{0.4}
\scalebox{0.95}{
  \begin{tabular}{ccc}
    \includegraphics[trim={0cm 0cm 16cm 0cm},clip, width=0.33\columnwidth]{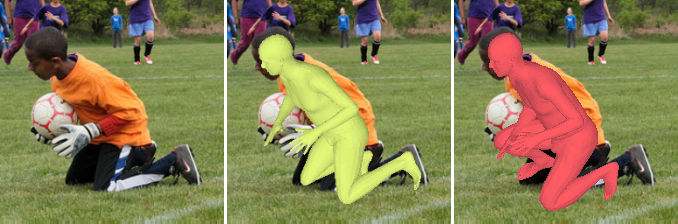} &
    \includegraphics[trim={16cm 0cm 0cm 0cm},clip, width=0.33\columnwidth]{figures/coco-spin-vs-ours/00095.png} &
    \includegraphics[trim={8cm 0cm 8cm 0cm},clip, width=0.33\columnwidth]{figures/coco-spin-vs-ours/00095.png} \\
    \includegraphics[trim={0cm 0cm 16cm 0cm},clip, width=0.33\columnwidth]{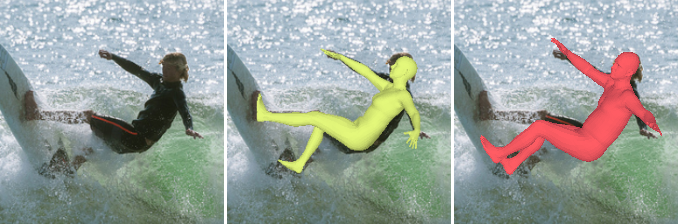} &
    \includegraphics[trim={16cm 0cm 0cm 0cm},clip, width=0.33\columnwidth]{figures/coco-spin-vs-ours/00928.png} &
    \includegraphics[trim={8cm 0cm 8cm 0cm},clip, width=0.33\columnwidth]{figures/coco-spin-vs-ours/00928.png} \\
    \includegraphics[trim={0cm 0cm 16cm 0cm},clip, width=0.33\columnwidth]{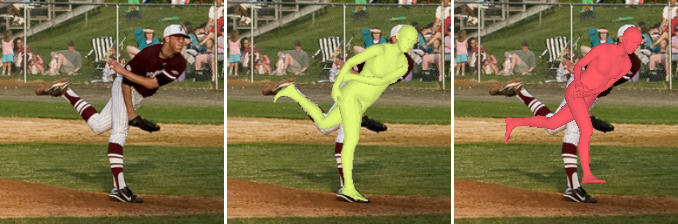} &
    \includegraphics[trim={16cm 0cm 0cm 0cm},clip, width=0.33\columnwidth]{figures/coco-spin-vs-ours/00257.png} &
    \includegraphics[trim={8cm 0cm 8cm 0cm},clip, width=0.33\columnwidth]{figures/coco-spin-vs-ours/00257.png} \\
    Input Image & SPIN~\cite{kolotouros2019spin} & Ours \\
  \end{tabular}
}
\caption{Qualitative comparison with the state-of-the-art method SPIN~\cite{kolotouros2019spin}. While mesh predictions of SPIN~\cite{kolotouros2019spin} are plausible, they do not align well with the image content, especially around the hand and feet regions. In contrast, our method yields accurate meshes with better alignment as it directly estimates body mesh from accurate 3D keypoints.\vspace{-5mm}}
\label{fig:teaser}
\end{figure}

The estimation of a human body mesh from a single RGB image is of great interest for numerous practical applications. The state-of-the-art methods~\cite{hmrKanazawa18, guler2019holo, kolotouros2019spin, choutas2020expose} in this area use deep neural networks with a fully-connected output layer, and directly regress the parameters of a parametric mesh model from the input image. While the performance of these methods has improved significantly, learning a mapping between images and mesh parameters in this way is highly non-linear. Therefore, these methods often suffer from low localization accuracy.  Specifically, while these methods estimate parameters that are plausible, the resulting meshes are often misaligned with the visual content, in particular, the wrists and feet regions (See Fig.~\ref{fig:teaser}). Additionally, these methods require a large number of images annotated with ground-truth meshes which is very hard to acquire specifically in unconstrained scenes. 

On the other hand, recent methods for 3D keypoint regression~\cite{sun18integeral, iqbal2020learning, sarandi2020metric} can accurately localize body keypoints with their 2D projections aligning well with the image content. Instead of learning a direct mapping between input images and 3D coordinates~\cite{zhou2016deep,tekin2016structured,popa2017CVPRmultitask,sun2017compositional}, these methods first estimate an intermediate volumetric~\cite{pavlakos2017volumetric, sun18integeral, luvizon20182d, sarandi2020metric} or heatmap-like~\cite{iqbal2020learning} representation, and then recover the 3D coordinates from them. This results in better 3D keypoint localization as better correspondences can be built between spatial image locations and the output 3D representation through fully-convolutional neural networks. Motivated by this, some recent works for mesh estimation also encode mesh vertex coordinates in heatmap-like representation~\cite{moon2020i2l}. While they show impressive performance, it comes at the cost of requiring large amounts of images annotated with body pose and shape labels. 

The ground-truth shape annotations used by the methods for body mesh estimation~\cite{hmrKanazawa18, kolotouros2019spin, choutas2020expose} are usually obtained using the seminal approach MOSH~\cite{Loper:SIGASIA:2014}. MOSH~\cite{Loper:SIGASIA:2014} shows that a sparse set of 3D marker locations on the human body are sufficient to capture body shape and soft-tissue deformations. Motivated by this, in this work, we propose to harness the superior localization ability of recent keypoint regressors~\cite{pavlakos2017volumetric, sun18integeral, luvizon20182d, sarandi2020metric,iqbal2020learning} and reconstruct full human body mesh from the regressed 3D keypoint positions only. A method with this capability offers two main advantages: 1) As compared to the traditional regression based methods~\cite{hmrKanazawa18, kolotouros2019spin}, the mesh estimates will be more accurate and align better with visual clues since the 3D keypoints can be localized more accurately from images~\cite{pavlakos2017volumetric, sun18integeral, luvizon20182d, sarandi2020metric, iqbal2020learning}. 2) It does not require images paired with ground-truth shape labels which are very hard to acquire. 

Some existing works propose solutions in this direction but formulate the problem as an optimization framework where the parameters of a parametric body model (\eg, SMPL~\cite{SMPL:2015}) are optimized to match the articulation of a regressed 3D pose~\cite{mehta2017vnect,monototalcapture2019}. Some other methods instead use 2D positions, but can also be extended to 3D~\cite{bogo2016keep,SMPL-X:2019,song2020human}. Optimization-based methods are, however, prone to local-minimas and are time consuming, in particular, when a parametric mesh model with large number of vertices (\eg SMPL has 6890 vertices) has to be optimized. In contrast, in this work, we propose an analytical solution using a set of straightforward geometrical operations that are not prone to local-minimas and have negligible computational cost while showing better performance.

We enable analytical mesh articulation by learning a 3D keypoint regressor that provides 3D positions of a sufficient number ($K{=}26$) of keypoints to accurately capture orientation of most body parts. The main challenge to learn such a regressor is to have ground-truth 3D annotations as the existing datasets do not provide annotations for enough keypoints. We show that such a regressor can be learned using synthetic and/or weakly-labeled data. For this, we build on the recent progress in weakly-supervised 3D keypoint learning~\cite{rohdin2018multiview, kocabas2019epipolar, iqbal2020learning}, and train our keypoint regressor using a combination of fully- and weakly-labeled data. The keypoints that do not have labeled 3D annotations are learned using weakly-labeled data in the form of unlabeled multi-view images along with a collection of images annotated with 2D positions only. 

Given the estimated 3D keypoint positions, we then present KAMA, which is an analytical method for Keypoint Aware Mesh Articulation. It uses a set of straightforward geometrical operations to articulate a canonical mesh using the regressed 3D keypoint positions. While KAMA already achieves state-of-the-art results, we further show that the meshes can be refined further by using a simple first-order optimization that removes the discrepancies between the regressed keypoints and articulated mesh. As shown in Fig.~\ref{fig:teaser}, our approach offers accurate mesh fitting and significantly better alignment as compared to the traditional regression based state-of-the-art method~\cite{kolotouros2019spin}. We evaluate our proposed approach on the challenging 3DPW and  Human3.6M  datasets where it achieves state-of-the-art results.  

\section{Related Work}

In the following, we discuss existing methods for 3D keypoint regression and body mesh recovery.

\noindent\textbf{3D Keypoint Regression}: 
These methods regress 3D keypoint positions from an RGB image~\cite{LiC14,Sijin2015iccv,zhou2016deep,tekin2016structured,popa2017CVPRmultitask,pavlakos2017volumetric,sun2017compositional, tekin2017fuse, zhou2017weakly, tome2017lifting, dabral18SFM, pavlakos2017volumetric, pavlakos2018ordinal, sun18integeral} or a 2D pose~\cite{martinez2017simple, Moreno_arxiv2016, chen2017matching,iqabl2018dual,hossain2018exploiting,RonchiAEP18} as input. Recently, this is achieved by training a deep neural network using ground-truth 3D pose annotations. Earlier methods regress 3D keypoints using holistic regression with a fully-connected output layer~\cite{LiC14, Sijin2015iccv, zhou2016deep, tekin2016structured, popa2017CVPRmultitask,tekin2017fuse}. More recent methods, however, adopt fully-convolutional networks to produce volumetric~\cite{pavlakos2017volumetric, sun18integeral, luvizon20182d, sarandi2020metric} or heatmap-like~\cite{zhou2019hemlets,iqbal2020learning} representations. This enables better correspondence between input image and the output 3D pose representation, and therefore, leads to higher localization accuracy. Since the acquisition of ground-truth 3D data is very hard, many recent works try to learn 3D keypoint regressors  in semi~\cite{rohdin2018geometry,rohdin2018multiview, yao2019monet, wu2016eccv, li2019boosting, pavllo2019temporal} and weakly~\cite{novotny2019C3DPO, wang2019nrsfm, drove2018can3d, wandt2019repnet, chen2019unsupervised, pavlakos2017harvesting, kocabas2019epipolar, iqbal2020learning, kundu2020self, mitra2020multiview} supervised ways. In this work, we build on the advances of these methods to learn additional 3D keypoints required for our method. 

\begin{figure*}[t]
\includegraphics[trim={0.0cm 4.7cm 0.0cm 0},clip,scale=1]{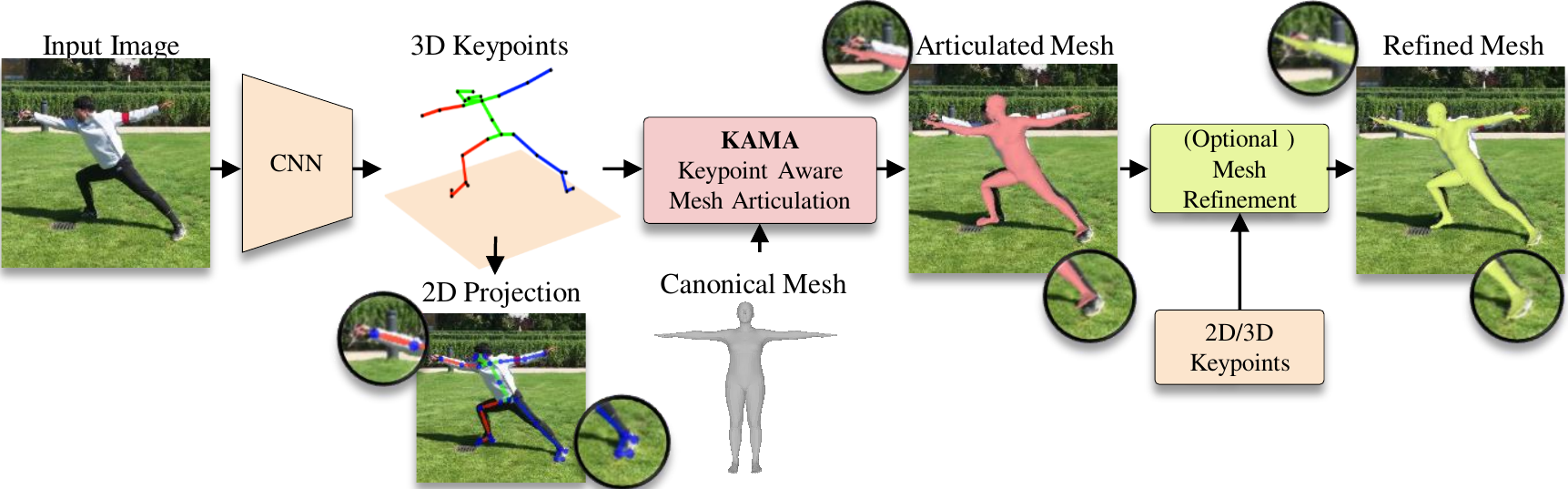}
\caption{\textbf{Overview:} Given an RGB image as input, we use a 3D keypoint regressor that produces absolute 3D positions of 26 body keypoints. We use the estimated 3D positions to articulate the SMPL body model using a set of geometrical transformation (Sec.~\ref{sec:mesh-recovery}). While KAMA already provides very good mesh reconstruction, the estimated mesh can be (optionally) improved further using a very simple first order optimization that minimizes the discrepancies between the articulated mesh and the regressed 2D and 3D keypoints (Sec.~\ref{sec:refinement-and-shape}). Our mesh estimates align much better with the image content as compared to state-of-the-art methods.\vspace{-5mm}} 
\label{fig:overview}
\end{figure*}

\noindent{\textbf{Body Mesh Recovery}}: These methods estimate body pose as well as its shape from RGB images. Most of the recent works adopt deep neural networks and directly regress the parameters of a parametric body model, SMPL~\cite{SMPL:2015}, from images~\cite{hmrKanazawa18, pavlakos2018humanshape, kolotouros2019spin, guler2019holo, pavlakos2019texture, Rong_2019_ICCV, choutas2020expose, kocabas2019vibe, sun2019human, joo2020eft, kundu2020mesh}. However, learning this non-linear mapping is very hard and often results in meshes that do not align very well with image content. The approaches~\cite{kolotouros2019convolutional, moon2020i2l} try to alleviate this problem by predicting the vertex coordinates from image features. While it leads to better predictions, the main limitations of these methods is that they rely heavily on ground-truth body shape annotations which are very hard to acquire. 

Other works try to decompose the problem into stages. They first estimate 2D and/or 3D keypoints from images and then estimate the mesh parameters using optimization based methods~\cite{Akhter:CVPR:2015,bogo2016keep, SMPL-X:2019, joo2018total, monototalcapture2019, song2020human} or use graph CNN to directly reconstruct the mesh~\cite{choi2020pose}.  These methods rely on large collection of motion capture data, \eg, AMASS~\cite{AMASS:ICCV:2019}, to learn strong body pose prior. The optimization based methods are, however, prone to local-minimas due to 2D-3D depth ambiguities and require careful initialization for optimal solutions. Also, they can be  computationally very intensive due to their iterative nature, in particular, when a body mesh with a large number of vertices has to optimized. Our work also falls into this category in that we first estimate the 3D body keypoints and then reconstruct the body mesh. However, we propose an analytical solution to articulate a canonical mesh using the estimated 3D keypoints and a set of geometric operations. Our proposed approach is neither expensive, nor it is prone to local minimas.

Some methods adopt a hybrid approach by utilizing a regression followed by optimization strategy. The approaches~\cite{guler2019holo, zanfir2020weakly} train regressors that produce several pose representations in addition to the parameters of SMPL. The additional representations are then used in an optimization framework to refine the initial SMPL predictions. This strategy is, however, extremely data hungry. In addition to the labels for body pose and shape, they also require segmented part labels and DensePose~\cite{Guler2018DensePose} annotations. Our approach can also benefit from this hybrid-strategy as we will show in the experiments. However, in contrast to these methods, our approach does not require 3D mesh annotations or any kind of additional labels such as part segmentation or DensePose. Yet, it outperforms these methods by a large margin.

\section{Method}

Our goal is to reconstruct the full 3D body mesh $\mathbf{M}$ from an RGB image $\mathbf{I}$ of a pre-localized person. We do this via a set of 3D body keypoints that are obtained using a learned keypoint regressor. In the following, we first describe our approach for 3D keypoint regression (Sec.~\ref{sec:keypoint-regression}) and then present our proposed method for body mesh articulation from the regressed keypoints (Sec.~\ref{sec:mesh-recovery}). In (Sec.~\ref{sec:refinement-and-shape}), we show that the estimated meshes can be refined further using a simple optimization objective and body mesh priors. An overview of the proposed approach can be seen in Fig.~\ref{fig:overview}.

\subsection{3D Keypoint Regression}
\label{sec:keypoint-regression}
Our goal is to learn a keypoint regressor $\mathcal{F}(\mathbf{I})$, in the form of a deep neural network, that takes an image $\mathbf{I}$ as input and produces the 3D positions $\mathbf{X}{=}{\{\mathbf{x}_k\}}_{k \in K}$ of $K$ body keypoints. Since we aim to use the keypoints to articulate a canonical mesh, the number of keypoints should be sufficient to obtain finer details about body mesh such as the head and feet orientation. The commonly used 17 keypoints are, however, insufficient for this purpose. For example, the 3D head pose (yaw, pitch, roll) in the mesh cannot be fully determined by the 3D positions of the neck and top-of-the-head locations only. We need additional 3D keypoints on the face to describe the full head pose. Therefore, in this work, we learn to regress $K{=}26$ body keypoints including, eyes, ears, nose, small and big toes, and heels, in addition to the other commonly used body keypoints. One main challenge to learn such a regressor is that the existing datasets for 3D human pose, such as the Human3.6M~\cite{h36m_pami} and MPII-INF-3DHP~\cite{mono20173dhp}, do not provide ground-truth annotations for the additional keypoints. To this end, we adopt the recent work of Iqbal~\etal~\cite{iqbal2020learning} that trains the 3D regressor using weakly-labeled data through multi-view consistency and 2D pose labels.  During training, we supervise the keypoints that have ground-truth 3D annotations using fully-supervised losses and train the remaining 9 keypoints (eyes, nose, ears, toes, and heels) using using weakly-supervised losses via multi-view consistency as done in~\cite{iqbal2020learning}. Such a joint fully and weakly-supervised training strategy allows us to train a 3D pose regressor with all $K{=}26$ keypoints given 2D annotations for all keypoints in one dataset (\ie, MS-COCO~\cite{lin2014microsoft,cao2019openpose}), and 3D annotations for some keypoints along with multi-view images from another dataset (\ie Human3.6M~\cite{h36m_pami}).

There are two additional useful properties of the keypoint regressor trained using~\cite{iqbal2020learning}. First, it provides absolute 3D positions of the keypoints. Therefore, we recover the body mesh in the absolute camera space and use perspective-projection to project it onto the image plane. Second, it reconstructs 3D keypoints using a 2.5D heatmap representation~\cite{iqbal2018hand}, which yields 3D keypoints that are well-aligned with the image content. An example of our estimated 3D keypoints can be seen in Fig.~\ref{fig:overview}. We refer the reader to \cite{iqbal2020learning} for further details about training the regressor.

\subsection{KAMA: Keypoint Aware Mesh Articulation}
\label{sec:mesh-recovery}

Given the regressed 3D keypoints $\mathbf{X}$ from the previous section, our goal is to articulate a canonical mesh such that it matches the pose of the person. We encode the body mesh using the Skinned Multi-Person Linear (SMPL) model~\cite{SMPL:2015}. SMPL represents the body mesh using a linear function $M(\theta, \beta)$  that takes as input the pose parameters $\theta \in \mathbb{R}^{24\times3}$ and the shape parameters $\beta \in \mathbb{R}^{10}$ and produces an articulated triangle mesh $\mathbf{M} \in \mathbb{R}^{V{\times}3}$ with $V{=}6980$ vertices. The pose parameters $\theta$ consist of local 3D-rotation matrices, in axis-angle format, corresponding to each joint in the pre-defined kinematic structure $\epsilon$ of the human body.
In the following, we estimate the pose parameters $\theta$ of the mesh from the regressed 3D keypoints using a set of geometrical transformations.
We use a simple procedure that is fully analytic and the computational cost is completely negligible.

\subsubsection{Keypoint Rotations from 3D Positions}
\label{sec:articulation}
Let ${\mathbf{\bar{M}}}$ be the body mesh in the canonical pose and ${\mathbf{\bar{X}}}{=}\mathbf{W\bar{M}}{=}{\{\mathbf{\bar{x}}_k\}}_{k \in K}$ be the 3D keypoint positions in the canonical pose. Here $\mathbf{W}\in\mathbb{R}^{K \times V}$ is a learned weight matrix that defines the contribution of every vertex to the keypoints. Our goal is to use $\V{X}$ and $\V{\bar{X}}$ to calculate a set of rotations  $\hat{\theta}{=}\{\mathbf{\theta}_k\}_{k \in K}$ such that the mesh $\V{\hat{M}}{=}M(\xi(\hat{\theta}), \beta{=}\V{0}^{1\times10})$ has an
articulation similar to that of the regressed keypoints $\V X$. Here the function $\xi(.)$ converts the order of rotation matrices from the 26-keypoint skeleton structure used for the keypoint regressor to the skeleton of SMPL which has 24 keypoints.   Following the definition of SMPL, we use axis-angle representation of the rotation matrices. We define $C(k)$ as the children keypoints of keypoint $k$ and $N(k)$ as the set of all keypoints adjacent to $k$, including $k$, as defined by the kinematic structure $\epsilon$. 

We apply three different rules to compute an initial estimation of the global rotations $\theta_k^g$ for every keypoint $k$: 1) For keypoints with one child we estimate rotation with ambiguous twist which is later compensated, 2) for keypoints with multiple children we estimate  rotation with the help of other connected joints that move rigidly with $k$, and 3) we assume no rotation for childless keypoints. These rules are summarized as follows:

\begin{equation}
\label{eq:rotation_rules}
\begin{split}
\theta_k^g &=
\begin{dcases}
     \alpha_1 (\V{{\bar{x}}}_{c(k)}{-}\V{\bar{x}}_k, \V{{x}}_{c(k)} {-} \V{x}_k)         & \text{if  } |C(k)| = 1 \\
     \alpha_2 (\V{\bar{X}}_k^\mathrm{N}, \V{{X}}_k^\mathrm{N}) & \text{if  } |C(k)| > 1 \\
     \V{0}^{1 \times 3}                                         & \text{otherwise}, \\
\end{dcases}
\end{split}
\end{equation}
where $\V{\bar{X}}_k^\mathrm{N} = \{\V{\bar{x}}_n\}_{_{n \in N(k)}}$ and $\V{{X}}_k^\mathrm{N}=\{\V{x}_n\}_{_{n \in N(k)}}$.

\noindent\textbf{Keypoints with one child:} For the keypoints with one child, we compute rotation as the angle applied to the vector perpendicular to the plane formed by the bones ${\V{{\bar{x}}}_{c(k)}{-}\bar{x}}_k$ and $\V{{x}}_{c(k)} {-} \V{x}_k$ in the canonical and estimated poses, respectively. $c(k)$ corresponds to the index of the child of keypoint $k$, and the function $\alpha_1(\V{v}_1,\V{v}_2)$ provides the rotation in axis-angle format as follows:
\begin{equation}
\alpha_1(\V{v}_1, \V{v}_2) = \mathrm{arccos}\big(\frac{{\V{v}}_1^T {\V{v}}_2}{||\V{v}_1|| \, ||\V{v}_2||}\big) \cdot \frac{\V{v}_1  \times \V{v}_2}{||\V{v}_1  \times \V{v}_2|| }, 
\label{eq:axis-angle}
\end{equation}
where the right part represents the axis of rotation and the left part corresponds to the angle of rotation. 

It is important to note that the rotation estimated in this way is inherently ambiguous as any arbitrary twist about the child vector can be applied without affecting the position of the child keypoint. We will remove such ambiguous twists after calculating rotations for all keypoints as explained later in this section.

\begin{figure}[t]
  \centering
    \setlength{\tabcolsep}{0.5pt}
\scalebox{0.95}{
  \begin{tabular}{ccc}
    \includegraphics[height=0.33\columnwidth]{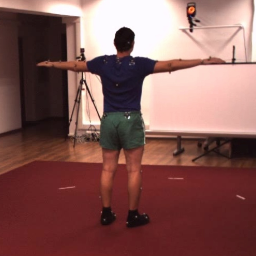} & 
    \includegraphics[height=0.33\columnwidth]{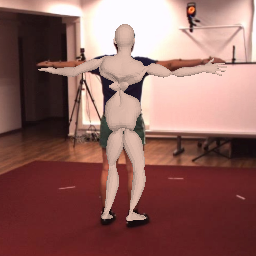} & 
    \includegraphics[height=0.33\columnwidth]{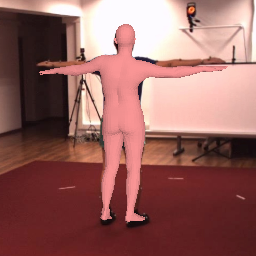} \\
  \end{tabular}
}
\caption{Illustration of ambiguous rotations for the keypoints with one child. Left: Input image. Middle: The articulated mesh has ambiguous twists around all keypoints with one child. Right: The articulated mesh after the removal of ambiguous twists. Both meshes are articulated using the same 3D keypoints.\vspace{-5mm}}
\label{fig:twist_removal}
\end{figure}

\noindent\textbf{Keypoints with multiple children:} 
For the keypoints with multiple children, we can estimate the keypoint rotation more precisely. Here we assume that all keypoints in $N(k)$ move rigidly, and estimate the rotation as a rigid rotation between $\V{\bar{X}}_k^\mathrm{N}$ and $\V{{X}}_k^\mathrm{N}$ as

\begin{equation}
\label{eq:multiple-children}
     \alpha_2 (\V{\bar{X}}_k^\mathrm{N}, \V{{X}}_k^\mathrm{N}) = \argmin_{\theta} \sum_{\substack{{\bar{\V{x}}_i \in \bar{\V{X}}_k^{\mathrm{N}}} \\ {\V{x}_i \in \V{X}_k^{\mathrm{N}}}}} \psi(\V{x}_i)(\phi(\theta,  \bar{\V{x}_i}) - \V{x}_i),
\end{equation}
where $\phi(\theta, \bar{\V{x}})$ represents rotating the vector $\V{x}$ with $\theta$ using Rodriguez formula, and $\psi(\V{x_i})$ corresponds to the detection confidence of keypoint $i$ as provided by the keypoint regressor $\mathcal{F(\V{I})}$. The eq.~\eqref{eq:multiple-children} can be solved easily in closed-form using singular value decomposition~\cite{procrustes1997}. Thanks to our 26-keypoint regressor, many of the keypoints (\ie, pelvis, neck, nose/face, ankles) fall into this category. 

\noindent\textbf{Global to local rotations:}
The rotations of the body joints as calculated above are the global rotations for each of them. However, to be able to use them in the function $M(\xi(\theta), \beta)$ to articulate the SMPL mesh, we convert them to local rotations as follows:
\begin{equation}
    \theta_k = \theta_{p(k)}^{g-1} \cdot \theta^g_k,
\end{equation}
where $p(k)$ is the index of the parent of keypoint $k$. The root keypoint has no parent so it remains unchanged. 

\noindent\textbf{Twist removal:} 
\label{sec:twist_removal}
Given the local rotations for all body joints, we need to remove unnecessary twists from the rotations of the joints with one child. A reasonable choice is to default to the twist from the canonical pose (which is zero by definition). This can be done via swing-after-twist decomposition~\cite{Dobrowolski2015SwingtwistDI}.
Specifically, we decompose the estimated local rotation into its swing and twist components, and then set the rotation as the swing component, while discarding the twist component. A comparison between the meshes before and after the twist removals can be seen in Fig~\ref{fig:twist_removal}.

\subsubsection{Scale and Translation Estimation}
So far, we have articulated the canonical mesh to match the pose of our regressed keypoints. However, it still lies at the origin and its global scale is unknown. Since our keypoint regressor provides absolute 3D pose including approximate bone length scales, we calculate the global translation  $\V{t} \in \mathbb{R}^3$ and global scale $s \in \mathbb{R}$ for the articulated mesh using Procrustes analysis~\cite{procrustes1997} between the keypoints of the mesh and regressed keypoints: 
\begin{equation}
    \hat{s}, \hat{\V{t}} = \underset{s, \V{t}}{\mathrm{argmin}} ||\V{W}(s\V{\hat{M}}{+}\V{t}) - \V{X}||^2_2.
    \label{eq:scale_and_translation}
\end{equation}
where $\V{\hat{M}}{=}M(\xi(\hat{\theta}), \beta{=}\V{0}^{1\times10})$ is the articulated mesh using the estimated rotations. This gives us our final articulated mesh in the absolute camera coordinate system as $\V{M}{=}\hat{s}\V{\hat{M}}{+}\hat{\V{t}}$.
We use perspective-projections to project the resulting mesh onto the image plane.

\begin{figure*}[t]
  \centering
    \setlength{\tabcolsep}{2pt}
    \renewcommand{\arraystretch}{0}
\scalebox{0.95}{
  \begin{tabular}{c|c}
    \includegraphics[width=0.49\textwidth]{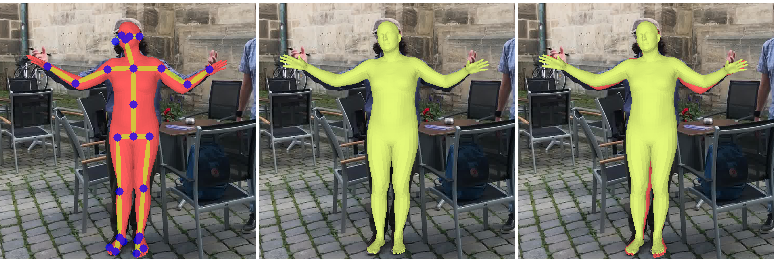} &
        \includegraphics[width=0.49\textwidth]{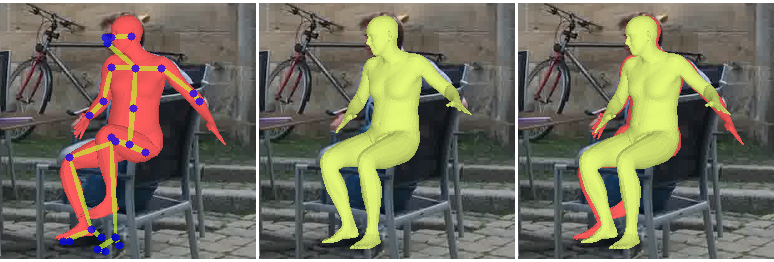} \\
    \includegraphics[width=0.49\textwidth]{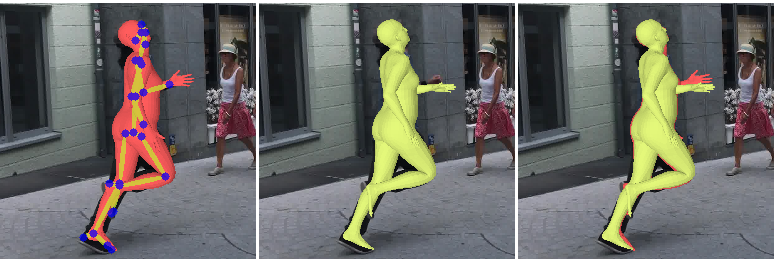} &
        \includegraphics[width=0.49\textwidth]{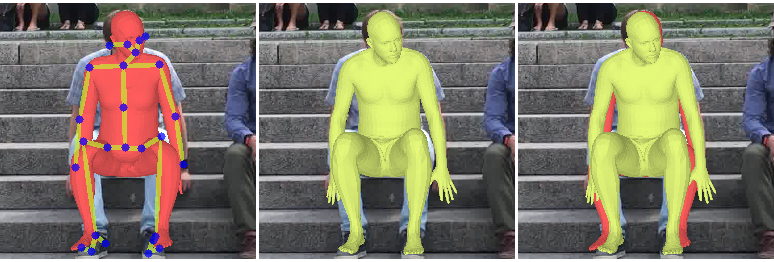} \\
    \begin{minipage}{.16\textwidth}\small\centering (a) KAMA\end{minipage}%
    \begin{minipage}{.16\textwidth}\small\centering (b) KAMA + refine\end{minipage}%
    \begin{minipage}{.16\textwidth}\small\centering (c) comparison\end{minipage}%
    & 
    \begin{minipage}{.16\textwidth}\small\centering (a) KAMA\end{minipage}%
    \begin{minipage}{.16\textwidth}\small\centering (b) KAMA + refine\end{minipage}%
    \begin{minipage}{.16\textwidth}\small\centering (c) comparison\end{minipage}\\
  \end{tabular}
 
}
\caption{(a) Articulated body meshes obtained using our proposed approach KAMA (Sec.~\ref{sec:mesh-recovery}). (b) Meshes obtained after pose and shape refinement  (Sec.~\ref{sec:refinement-and-shape}). (c) Comparison between a \& c. While KAMA already provides very good mesh estimates, they sometimes can have errors due to missing twist information, errors in the regressed keypoints, error propagation, occlusions, and etc. Such errors can be fixed using a simple optimization objective consisting of body pose and shape priors.\vspace{-5mm}}
\label{fig:kama-vs-opt}
\end{figure*}

\subsection{Pose Refinement and Shape Estimation}
\label{sec:refinement-and-shape}

While our approach for mesh articulation using keypoints already provides state-of-the-art mesh estimates, as we will show later in our experiments, there are a few issues that can be addressed further. First, our method resorts to canonical twist for the keypoints with single children which is not the most optimal choice. Second, the regressed keypoints do not exactly match with the skeleton structure of SMPL. For example, in contrast to SMPL, the regressor does not provide any keypoints on the collar bones. Depending on the 2D annotations, there can be other subtle differences between the estimated keypoints and the keypoints in the canonical mesh. Also, small errors in one keypoint can propagate to the entire mesh. For example, an incorrect rotation for pelvis will impact all other keypoints and will result in mesh keypoints that are very different from the regressed keypoints. Lastly, we also need to estimate the shape parameters $\beta$ of SMPL to fully capture the body details. 
To this end, we build on~\cite{bogo2016keep} and remove such discrepancies by using body pose and shape priors in an energy minimization formulation that further refines the pose parameters $\theta$, shape $\beta$, global translation $\V t$, and global scale $s$:
\begin{equation}
    \hat{\theta}, \hat{\beta}, \hat{\V t}, \hat{s} = \underset{\theta, \beta, \V t, s}{\mathrm{argmin}}~ \mathcal{L}(\theta, \beta, \V t, s), 
    \label{eq:optimization}
\end{equation}
where $\mathcal{L}(\theta, \beta, \V t, s)$ consists of four errors terms
\begin{equation}
    \mathcal{L}(\theta, \beta, \V{t}, s) = \mathcal{L}_{\text{2D}} + \omega_1 \mathcal{L}_{\text{3D}} + \omega_3 \mathcal{L}_{\theta} + \omega_2 \mathcal{L}_{\beta}. 
    \label{eq:total_error}
\end{equation}

The error term $\mathcal{L}_{\text{2D}}$ is the reprojection error. It measures the discrepancies between the 2D keypoints provided by the regressor and the projection of the mesh skeleton $\hat{\V{X}}{=}\{\hat{\V{x}}_k\}_{k \in K}{=}\V W \V (s\V{M}+ \V t)$:
\begin{equation}
 \mathcal{L}_{\text{2D}} = \sum_k \psi(\V{x}_k) ||P(\V{K},\V{x}_k) - P(\V{K},\hat{\V{x}}_k)||_2^2,
\end{equation}
where $\V{K}$ is the intrinsic camera matrix, $P(.,.)$ represents projection on the image plane, $\psi(\V{x}_k)$ corresponds to the detection score of the keypoint $k$. $\mathcal{L}_\text{3D}$ measures the difference between predicted 3D position and the 3D mesh skeleton:
\begin{equation}
 \mathcal{L}_{\text{3D}} = \sum_k \psi(\V{x}_k) ||\V{x}_k - \hat{\V{x}}_k||_2^2.
\end{equation}

The error terms $\mathcal{L}_\theta$ and $L_\beta$ correspond to the pose prior and shape prior terms as defined in~\cite{bogo2016keep}, respectively. Specifically,  $\mathcal{L}_\theta$ favors plausible pose parameters $\theta$. In our case, it helps in recovering the optimal twist for keypoints with one child and in reducing the ambiguities due to missing keypoints and differences in the skeleton structures. The term $\mathcal{L}_\beta$ is a regularization for parameters $\beta$ such that the optimized shape is not distant from the mean shape. 

For the optimization, we use the values of $\theta$, $s$, and $\V{t}$ from KAMA as initialization and use Adam as the optimizer. Since we start from a very good initialization, we found that the optimization converges within 100 iterations without the need of a multi-stage optimization strategy as required by prior works~\cite{bogo2016keep, SMPL-X:2019}. Some examples of mesh estimates before and after the refinement can be seen in Fig.~\ref{fig:kama-vs-opt}

\section{Implementation Details}
We follow~\cite{iqbal2020learning} and use HRNet-w32~\cite{SunXLW19} as the keypoint regressor. We empirically choose $\omega_1=500$, and adopt $\omega_2 =  4.78$ and  $\omega_{3} = 5$ from \cite{bogo2016keep, kolotouros2019spin}. We use the publicly available implementation of SMPL provided by~\cite{kolotouros2019spin}. The linear regressor $\V{W}$ in this implementation allows to extract 54 keypoints from the mesh vertices. We choose 26 keypoints that are closest to the 2D annotations used by our keypoint regressor. Note that these keypoints do not exactly overlap with the native 24 keypoints of SMPL, but are sufficient to calculate enough rotation matrices in $\theta \in \mathbb{R}^{24\times3}$ to capture the full body pose. The rotation matrices that cannot be estimated (\ie, collar-bones, spine-1, spine-3, and hands) are assigned zeros in~\eqref{eq:rotation_rules}, but optimized in~\eqref{eq:optimization}. 
\section{Experiments}
In this section, we evaluate the performance of the proposed approach in detail and also compare it with the state-of-the-art methods for human pose and shape estimation. 

\subsection{Datasets}
\noindent\textbf{Human3.6M}~\cite{h36m_pami}: We follow the standard protocol~\cite{hmrKanazawa18} and use five subjects (S1, S5, S6, S7, S8) for training and test on two subjects (S9 and S11) on the frontal camera view. 

\noindent\textbf{3D Poses in-the-Wild} (3DPW)~\cite{vonMarcard2018}: consists of 60 videos recorded in diverse environments. The ground-truth poses are obtained with the help of IMU sensors attached to the body. We follow the standard protocol and use its test-set for evaluation and do not train on this dataset. 

\noindent\textbf{RenderPeople}: This is a synthetic dataset with ground-truth annotations for all $K{=}26$ keypoints used in our method.  We used 10 characters from RenderPeople~\cite{renderpeople} dataset and generated 80k images under a variety of poses using CMU MoCap dataset~\cite{cmu_mocap} while using $\sim$100 outdoor HDRI image from HDRI Haven~\cite{hdrihaven} for lighting and backgrounds. We manually annotated the vertices corresponding to eyes, ears, nose, toes and heels for each character as they are not part of the rigged skeletons.

\noindent\textbf{MS-COCO} (COCO)~\cite{lin2014microsoft}: We use this dataset as the weakly-labeled set for the training of the 3D keypoint regressor. The dataset provides 2D annotations for 18 keypoints. A subset of the dataset was augmented by~\cite{cao2019openpose} with annotations for 3 additional keypoints on each foot. 

\subsection{Evaluation and Training Setting} We report Mean Per-Vertex Error (MPVE) and 3D reconstruction error in millimeters (mm) for all experiments. Following the standard practice \cite{hmrKanazawa18}, we extract 14 keypoints for evaluation from the recovered mesh using a pre-trained linear regressor.  

For 3DPW dataset, different methods use different datasets for training which include Human3.6M~\cite{h36m_pami}, MSCOCO~\cite{lin2014microsoft}, MPI-INF-3DHP~\cite{mono20173dhp}, MuCo-3DHP~\cite{mehta2018multi}
MPII~\cite{andriluka14cvpr}, LSP~\cite{johnson2010lsp}, UP~\cite{lassner2017unite}, SURREAL~\cite{varol17b} and etc. In this work, we only use Human3.6M, MSCOCO and the 80k synthetic images from RenderPeople dataset to train the model used for evaluation on 3DPW dataset. 

For evaluation on Human3.6M, we train only using Human3.6M and MSCOCO datasets. The ground-truth mesh annotations for Human3.6M are only available to a sparse set of researchers as the distribution has been  discontinued. Hence, we only report reconstruction error for Human3.6M. We also found that there are discrepancies between the 14 keypoints extracted using the keypoint regressor provided by~\cite{bogo2016keep} and the ground-truth marker locations from Human3.6M. This is not a problem when ground-truth mesh annotations are available as the consistent ground-truth keypoints can be extracted from the meshes. Since the mesh annotations are not available to us, we remove these discrepancies by training another linear regressor ($42{\times}42$ weight matrix) using training data, and apply it to the extracted 14 keypoints before evaluation (see supp. for details).

\begin{table}[t]
\scriptsize
\centering
\begin{tabularx}{1\columnwidth}{X|cc}
\toprule
\multirow{1}{*}{\bf Methods} & \bf MPVE & \bf Recon. Error \\
\midrule
\multicolumn{3}{c}{articulation using 3D keypoints \eqref{eq:rotation_rules}} \\
\midrule
% INFO: :MPJPE: 103.7087, PA-MPJPE: 64.3305, PVE: 124.7830, ACCEL: 34.6182, ACCEL_ERR: 35.0869,
KAMA w/o twist removal                & 124.8 & 64.0  \\
% INFO: :MPJPE: 93.0985, PA-MPJPE: 54.4790, PVE: 107.6955, ACCEL: 32.1310, ACCEL_ERR: 32.6055,
KAMA with twist removal                   & 107.7 & 54.5 \\
\midrule
\multicolumn{3}{c}{pose \& shape refinement using~\eqref{eq:optimization}} \\
\midrule
Initialization using~\eqref{eq:rotation_rules}   \\ 
% No initialization w. 500 iter, & 50.4 & 62.8   \\ 
% INFO: :MPJPE: 126.2510, PA-MPJPE: 87.8369, PVE: 152.5137, ACCEL: 101.6312, ACCEL_ERR: 101.8139,
\quad $\mathcal{L}_\text{2D}$                                                   & 152.5 & 87.8  \\
% INFO: :MPJPE: 99.5309, PA-MPJPE: 63.5302, PVE: 115.8364, ACCEL: 44.4877, ACCEL_ERR: 44.8209,
\quad $\mathcal{L}_\text{2D}$ +  $\mathcal{L}_\text{3D}$                        & 115.8 & 63.5 \\
% INFO: :MPJPE: 86.8907, PA-MPJPE: 53.0153, PVE: 100.3963, ACCEL: 33.3789, ACCEL_ERR: 33.7750,
\quad $\mathcal{L}_\text{2D}$ +  $\mathcal{L}_\text{3D}$ + $\mathcal{L}_\theta$ & 100.4 & 53.0 \\
% INFO: :MPJPE: 83.3756, PA-MPJPE: 51.1078, PVE: 96.9694, ACCEL: 32.5436, ACCEL_ERR: 32.9437,
\quad $\mathcal{L}_\text{2D}$ +  $\mathcal{L}_\text{3D}$ + $\mathcal{L}_\theta$  + $\mathcal{L}_\beta$   & 97.0 & 51.1 \\   
\midrule
% INFO: :MPJPE: 91.4518, PA-MPJPE: 56.5439, PVE: 106.6458, ACCEL: 36.5248, ACCEL_ERR: 36.9031,
No initialization                                   & 106.6      &  56.4  \\
% INFO: :MPJPE: 87.9616, PA-MPJPE: 55.8202, PVE: 100.0679, ACCEL: 30.7482, ACCEL_ERR: 31.2015,
SMPLify3D - initialization using mean pose          & 100.7 & 55.8   \\ 
% INFO: :MPJPE: 104.6564, PA-MPJPE: 69.2848, PVE: 115.0031, ACCEL: 51.1376, ACCEL_ERR: 51.4852,
SMPLify2D - initialzation using mean pose           &   115.0    &  69.3       \\
\midrule
\multicolumn{3}{c}{impact of 3D keypoint quality - using GT 3D keypoints} \\
\midrule
% INFO: :MPJPE: 40.3017, PA-MPJPE: 18.0123, PVE: 47.4320, ACCEL: 6.1747, ACCEL_ERR: 2.1684,
KAMA                &   47.4   &  18.0 \\
% INFO: :MPJPE: 38.0798, PA-MPJPE: 17.0378, PVE: 44.1859, ACCEL: 8.0008, ACCEL_ERR: 4.4862,
KAMA with refinement using~\eqref{eq:optimization}                &   44.1 & 17.0 \\
\bottomrule
\end{tabularx}
\caption{Impact of different components in the proposed approach.\vspace{-5mm}}
\label{tab:ablation}
\end{table}

\subsection{Ablation Study}
In Tab.~\ref{tab:ablation}, we evaluate all components of the proposed approach. We chose 3DPW datasets for all ablative studies as it represents more general in-the-wild scenarios, and also provides ground-truth mesh annotations. First, we evaluate the performance of our approach for mesh articulation (KAMA) using regressed 3D keypoints. If we do not remove ambiguous twists from the calculated rotations, as explained in Sec~\ref{sec:twist_removal}, the estimated meshes yield a MPVE and 3D reconstruction error of 124.8mm and 64.0mm, respectively. Removing the ambiguous twists results in a significant decrease in the error (124.8mm vs. 107.7mm and 64.0mm vs. 54.5mm) and shows the importance of this step. Note that ambiguous twists have higher impact on the MPVE since the body surface is impacted more with incorrect twist rotations as compared to body keypoint positions. We would like to emphasize that the errors of 107.7mm and 54.5mm are the state-of-the-art on 3DPW dataset, even though we only use mean shape values (\ie, $\beta=\V{0}^{1\times10})$) in KAMA. Thanks to eq.~\eqref{eq:scale_and_translation}, we can find an optimal global scale for the mesh without having to optimize beta.

Next, we evaluate the contributions of different error terms used in~\eqref{eq:total_error}. In all cases we initialize the body joint rotations using~\eqref{eq:rotation_rules} and body translation and scale using~\eqref{eq:scale_and_translation}. If we only optimize for the re-projection loss $\mathcal{L}_{\text{2D}}$, the errors increase significantly (from 107.7mm~to~152.5mm and 54.5mm~to~87.8mm) due to the well known 2D-3D ambiguities. Adding $\mathcal{L}_{\text{3D}}$ reduces the errors significantly (from 152.2~to~115.8mm and 87.8mm~to~63.5mm)  but remain higher than what we can achieve by using KAMA only. This is because optimizing 2D and 3D losses only is still susceptible to ambiguous twists. In KAMA, on the other hand, we explicitly handle the twist by either discarding it or estimating it with the help of adjacent keypoints. Enforcing body pose priors using $\mathcal{L}_{\theta}$ significantly reduces the errors. As compared to KAMA only, the MPVE and joint reconstruction errors are reduced from 107.7mm~to~100.4mm and 54.5mm~to~53.0mm, respectively.  As mentioned earlier, $\mathcal{L}_\theta$ encourages plausible poses. In our case, it helps to recover the twists of keypoints with single child and to reduce the ambiguities due to missing 3D keypoints and differences in the skeleton structures. Finally, adding $\mathcal{L}_\beta$ results in further decrease in the errors demonstrating the importance of optimal body shape parameters. 

To emphasize the usefulness of KAMA for optimization based method, we also evaluate the case when no initialization for $\theta$ is used.  Since we have the estimated 3D keypoints, we can obtain reasonable initial values for global scale $s$, global translation $\V{t}$ and the global orientation $\theta_0$ by calculating a rigid transformation between the regressed keypoints and skeleton of the canonical mesh using Procrustes analysis. We initialize  $\beta$ and $\theta$ with zeros. Optimizing~\eqref{eq:optimization} without a good initialization results in a 3D error of 56.4mm which is significantly higher than the case when KAMA is used as the initialization (51.1mm),  demonstrating the importance of KAMA and accurate initialization. 

\begin{table}[t]
\footnotesize
\centering
\begin{tabularx}{1.0\columnwidth}{Xc|c|cc}
\toprule
\multirow{2}{*}{Methods}                    &           & Mesh                  & \multirow{2}{*}{MPVE}    & Recon.  \\
                                            &           & Supervision           &                          & Error  \\
\midrule
SMPLify \cite{bogo2016keep}*                 & ECCV'16   & \textcolor{green}{N}  &   -     &  106.1 \\
HMR                                         & CVPR'18   & \textcolor{red}{Y}    & 161.0   & 81.3  \\
Kundu \etal \cite{kundu2020mesh}            & ECCV'20   & \textcolor{green}{N}  &  -      & 78.2  \\
ExPose \cite{choutas2020expose}             & ECCV'20   & \textcolor{red}{Y}    & -       & 60.7  \\
Rong~\etal~\cite{Rong_2019_ICCV}            & CVPR'19   & \textcolor{red}{Y}    & 152.9   & -     \\
SPIN   \cite{kolotouros2019spin}            & ICCV'19   & \textcolor{red}{Y}    & 112.8   & 59.2  \\
Pose2Mesh \cite{choi2020pose}               & ECCV'20   & \textcolor{red}{Y}    & -       & 58.9  \\
I2L-Mesh  \cite{moon2020i2l}                & ECCV'20   & \textcolor{red}{Y}    & 110.1   & 58.6  \\
Zanfir \etal \cite{zanfir2020weakly}        & ECCV'20   & \textcolor{red}{Y}    &   -     & 57.1  \\
Song \etal \cite{song2020human}*             & ECCV'20   & \textcolor{green}{N}  & -       &  55.0 \\
\midrule
KAMA (ours)                                        &           & \textcolor{green}{N}    & \bf 107.7 & \bf 54.5  \\
\multicolumn{2}{l|}{KAMA w. refinement*}                 & \textcolor{green}{N}   & \bf 97.0 & \bf 51.1  \\
\bottomrule
\end{tabularx}
\caption{Comparison with state-of-the-art methods on \textbf{3DPW} dataset. *optimization-based methods\vspace{-5mm}}
\label{tab:tab:sota_3dpw}
\end{table}

In order to further evaluate the impact of initialization, we also implement a 3D version of SMPLify~\cite{bogo2016keep} which initializes the pose parameters $\theta$ with the mean body pose. We initialize the global orientation using the rotation of the $\textrm{pelvis}$ keypoint obtained using~\eqref{eq:rotation_rules} and global scale and translation using~\eqref{eq:scale_and_translation}. This results in a MPVE and 3D reconstruction error of 100.7mm and 55.8mm, respectively, that are significantly higher than the case when predictions from KAMA are used for initialization, demonstrating that KAMA can be used as a very good initialization approach. In fact, KAMA without any optimization achieves better 3D reconstruction error than SMPLify3D (54.5mm vs 55.8). Finally, for completeness, we also implement a 2D version of SMPLify by removing $L_{3D}$ from the objective while keeping all other error terms and initialization as SMPLify3D. The errors increase significantly showing that the 3D keypoints are important for accurately reconstruction. 

Finally, we also evaluate the impact of 3D keypoint accuracy on body mesh reconstruction using KAMA. For this, we extracted $26$ keypoints from the ground-truth meshes, and used KAMA to reconstruct body mesh. This setting serves as an upper-bound for KAMA. Using ground-truth 3D keypoints significantly decreases the errors showing that the performance can be improved further by using a more accurate 3D keypoint regressor. Notably, the difference between KAMA and KAMA-with-refinement decreases which indicates that the refinement step may not be required with more accurate 3D keypoints.

\begin{figure*}[hbt!]
  \centering
    \setlength{\tabcolsep}{1pt}
    \renewcommand{\arraystretch}{0.1}
\scalebox{0.95}{
  \begin{tabular}{ccc}
    \includegraphics[trim={9.1cm 0cm 0cm 0cm},clip, width=0.33\textwidth]{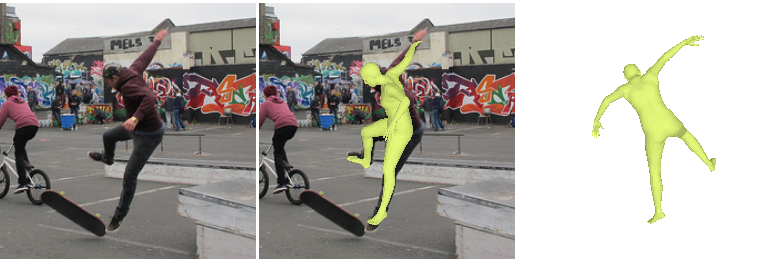} &
    \includegraphics[trim={9.1cm 0cm 0cm 0cm 0},clip,width=0.33\textwidth]{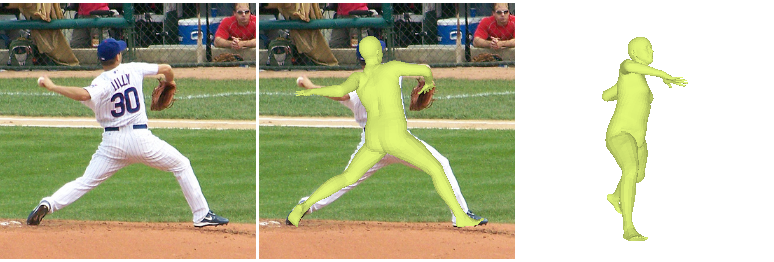} & 
    \includegraphics[trim={9.1cm 0cm 0cm 0cm 0},clip,width=0.33\textwidth]{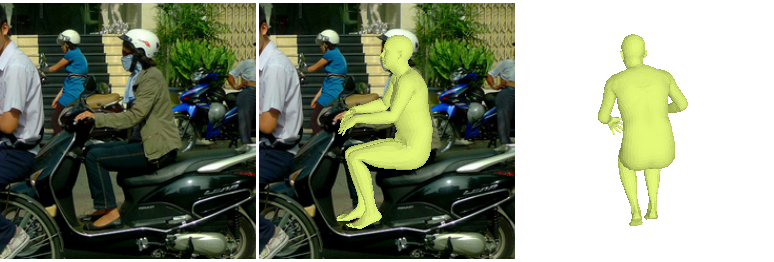} \\
        \includegraphics[trim={9.1cm 0cm 0cm 0cm},clip, width=0.33\textwidth]{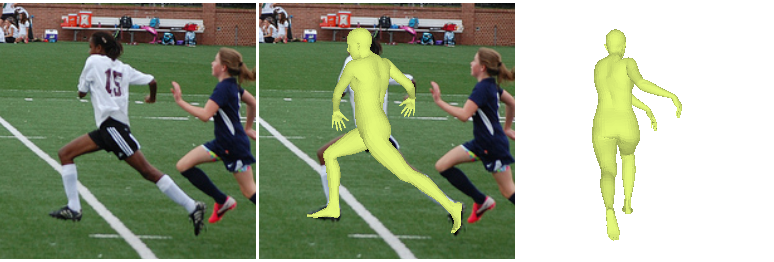} &
    \includegraphics[trim={9.1cm 0cm 0cm 0cm 0},clip,width=0.33\textwidth]{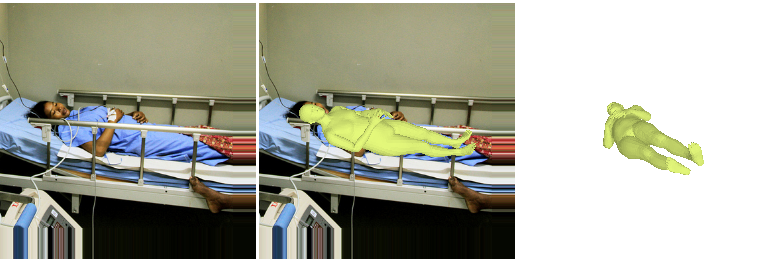} & 
    \includegraphics[trim={9.1cm 0cm 0cm 0cm 0},clip,width=0.33\textwidth]{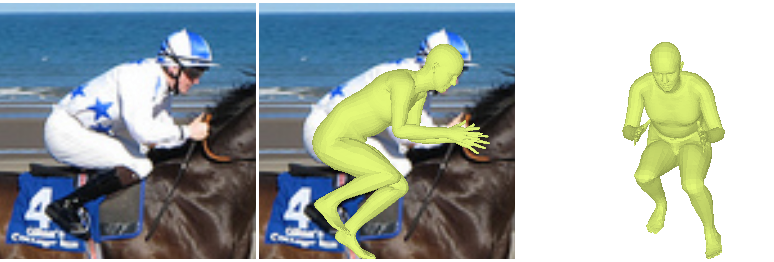} \\
    \includegraphics[trim={9.1cm 0cm 0cm 0cm},clip, width=0.33\textwidth]{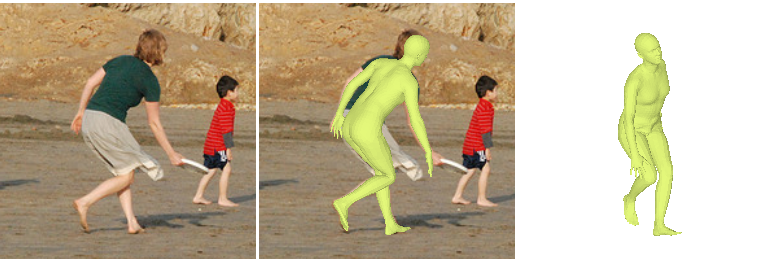} &
    \includegraphics[trim={9.1cm 0cm 0cm 0cm 0},clip,width=0.33\textwidth]{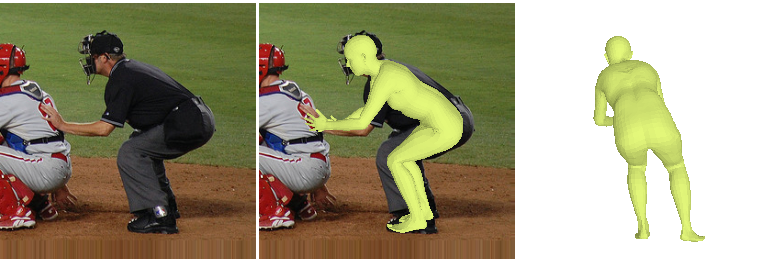} & 
    \includegraphics[trim={9.1cm 0cm 0cm 0cm 0},clip,width=0.33\textwidth]{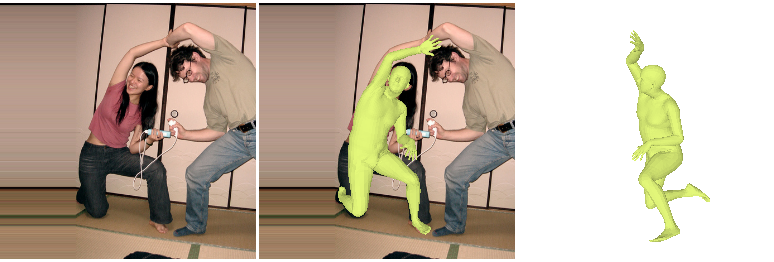} \\
  \end{tabular}
}
\caption{Some qualitative results from the validation set of COCO dataset.}
 \vspace{-4mm} 
\label{fig:qualitative}
\end{figure*}

\subsection{Comparison to the State-of-the-Art}
We compare the performance of our approach with the state-of-the-art on 3DPW and Human3.6M datasets. We only compare with methods that use static images as input and perform body mesh estimation. 

Tab.~\ref{tab:tab:sota_3dpw} compares our proposed method with the state-of-the-art on 3DPW dataset. We chose the best numbers reported in all papers. The MPVE for SPIN~\cite{kolotouros2019spin} and I2L-Mesh~\cite{moon2020i2l} are obtained using the publicly available source codes, whereas the MPVE for HMR is obtained from~\cite{Rong_2019_ICCV}. KAMA without any additional refinement already outperforms all state-of-the-art methods. Refining the mesh estimates using~\eqref{eq:optimization} further improves the results and sets a new state-of-the-art on 3DPW dataset. Note that the methods~\cite{hmrKanazawa18, kolotouros2019convolutional, kolotouros2019spin, pavlakos2019texture, guler2019holo, moon2020i2l, zanfir2020weakly} use images annotated with ground-truth mesh annotations, and the method~\cite{song2020human} relies on very strong pose priors learned from a massive corpus of MOSHed~\cite{Loper:SIGASIA:2014} motion capture data, AMASS~\cite{AMASS:ICCV:2019}.
KAMA, in contrast, does not require any mesh annotations and is trained using 3D keypoints supervision only, yet it outperforms them with a large margin. This demonstrates that a good keypoint regressor combined with KAMA can yield state-of-the-art mesh reconstruction without the need of hard-to-acquire body mesh annotations.  

\begin{table}[t]
\footnotesize
\centering
\begin{tabularx}{1\columnwidth}{Xc|c|c}
\toprule
\multirow{2}{*}{Methods}                    &           & Mesh                  &     Reconstruction  \\
                                            &           & Supervision           &       Error   \\

\midrule
SMPLify  \cite{bogo2016keep}*                            & ECCV'16  & \textcolor{green}{N} & 82.3 \\
SMPLify-X \cite{SMPL-X:2019}*                            & CVPR'19  & \textcolor{green}{N} & 75.9 \\
HMR  \cite{hmrKanazawa18}                               & CVPR'18  &  \textcolor{red}{Y} & 56.8\\
Song \etal \cite{song2020human}*                         & ECCV'20  & \textcolor{green}{N} & 56.4 \\
GraphCMR \cite{kolotouros2019convolutional}             & CVPR'19  & \textcolor{red}{Y} & 50.1\\
STRAPS \cite{akash2020synthetic}                        & BMVC'20  & \textcolor{green}{N} & 55.4 \\
Pose2Mesh \cite{choi2020pose}                           & ECCV'20  & \textcolor{red}{Y} & 47.0 \\
TexturePose \cite{pavlakos2019texture}                  & ICCV'19  & \textcolor{red}{Y} & 49.7 \\
Kundu \etal \cite{kundu2020mesh}                        & ECCV'20  & \textcolor{green}{N} & 48.1\\
HoloPose \cite{guler2019holo}                           & CVPR'19  & \textcolor{red}{Y} & 46.5 \\
DSD \cite{sun2019human}                                 & ICCV'19  & \textcolor{red}{Y} & 44.3 \\
% EFT \cite{joo2020eft}*                                  & ArXiv'20 & Y & 43.7 \\
%INFO: :MPJPE: 75.1860, PA-MPJPE: 47.1709, PVE: 0.0000, ACCEL: 204.5069, ACCEL_ERR: 87.6002,
I2L-MeshNet \cite{moon2020i2l}                          & ECCV'20 & \textcolor{red}{Y} &  41.7  \\
SPIN \etal  \cite{kolotouros2019spin}                   & ICCV'19 & \textcolor{red}{Y} & 41.1 \\
\midrule
Ours                                                    &  & \textcolor{green}{N} & \bf 41.5  \\
Ours w. refinement*                                      &  & \textcolor{green}{N} & \bf 40.2 \\
\bottomrule
\end{tabularx}
\caption{Comparison with the state-of-the-art methods on \textbf{Human3.6M} dataset. *optimization-based methods 
\vspace{-5mm}
}
\label{tab:sota_h36m}
\end{table}

Tab.~\ref{tab:sota_h36m} compares our proposed method with the state-of-the-art on Human3.6M dataset. On this dataset, KAMA without refinement performs on-par with SPIN~\cite{kolotouros2019spin} and I2L-MeshNet~\cite{moon2020i2l}. This is likely because of the limited diversity of Human3.6M where these methods can overfit using the full mesh annotations. Nonetheless, as before, refining the predictions of KAMA using~\eqref{eq:optimization} reduces the errors and results in state-of-the-art performance on Human3.6M. We would also like to emphasize that KAMA, unlike many other methods, provides meshes in absolute camera coordinates and uses perspective projection to project the meshes on to the images. This is in contrast to \eg, SPIN~\cite{kolotouros2019spin} and other methods that only predict root-relative meshes and use weak-perspective projection, hence, incur lower errors as compared to our method. 

Finally, in Fig.~\ref{fig:qualitative}, we provide some qualitative results of our approach on  in-the-wild images taken from MS-COCO~\cite{lin2014microsoft} dataset.\footnote{More qualitative results: \url{https://youtu.be/mPikZEIpUE0}}

\section{Conclusion}
In this work, we presented a novel approach for human mesh recovery from 3D keypoint positions only. To this end, we used a 3D keypoint regressor that is able to estimate 3D positions of 26 body keypoints. We then presented, KAMA, a 3D keypoint aware approach to articulate a canonical mesh using 3D keypoint positions and a set of simple geometrical operations. We then further improved the mesh estimates via a pose refinement and shape estimation approach. The resulting meshes are accurate and align well with image content. In contrast to existing methods, our approach does not require any 3D annotations and provides meshes in the absolute camera coordinate system. Yet, it achieves state-of-the-art results on the challenging 3DPW and Human3.6M datasets.

%\newpage

\balance
{\small
\bibliographystyle{ieee_fullname}
\bibliography{pose}
}
\clearpage

\end{document}